\documentclass[10pt,twocolumn,letterpaper]{article}

\usepackage{iccv}
\usepackage{times}
\usepackage{epsfig}
\usepackage{graphicx}
\usepackage{amsmath}
\usepackage{amssymb}

\usepackage{multirow}
\usepackage{csquotes}
\usepackage[ruled]{algorithm2e}
\usepackage[table]{xcolor}

\usepackage[breaklinks=true,bookmarks=false]{hyperref}

\iccvfinalcopy 

\def\iccvPaperID{****} 

\ificcvfinal\fi

\begin{document}

\title{Class-Incremental Learning using Diffusion Model for Distillation and Replay}

\author{Quentin Jodelet$^{1,2}$, Xin Liu$^{2}$, Yin Jun Phua$^{1}$, Tsuyoshi Murata$^{1,2}$ \\
{\small $^{1}$ Department of Computer Science, Tokyo Institute of Technology, Japan}\\
{\small$^{2}$ Artificial Intelligence Research Center, AIST, Japan}\\
{\small jodelet@net.c.titech.ac.jp,  xin.liu@aist.go.jp,  phua@c.titech.ac.jp,  murata@c.titech.ac.jp}
}

\maketitle
\ificcvfinal\thispagestyle{empty}\fi

\begin{abstract}
Class-incremental learning aims to learn new classes in an incremental fashion without forgetting the previously learned ones. Several research works have shown how additional data can be used by incremental models to help mitigate catastrophic forgetting. In this work, following the recent breakthrough in text-to-image generative models and their wide distribution, we propose the use of a pretrained Stable Diffusion model as a source of additional data for class-incremental learning. Compared to competitive methods that rely on external, often unlabeled, datasets of real images, our approach can generate synthetic samples belonging to the same classes as the previously encountered images. This allows us to use those additional data samples not only in the distillation loss but also for replay in the classification loss. Experiments on the competitive benchmarks CIFAR100, ImageNet-Subset, and ImageNet demonstrate how this new approach can be used to further improve the performance of state-of-the-art methods for class-incremental learning on large scale datasets.
\end{abstract}

\section{Introduction}
\begin{figure*}
\begin{center}
\resizebox{\textwidth}{!}{%
\includegraphics{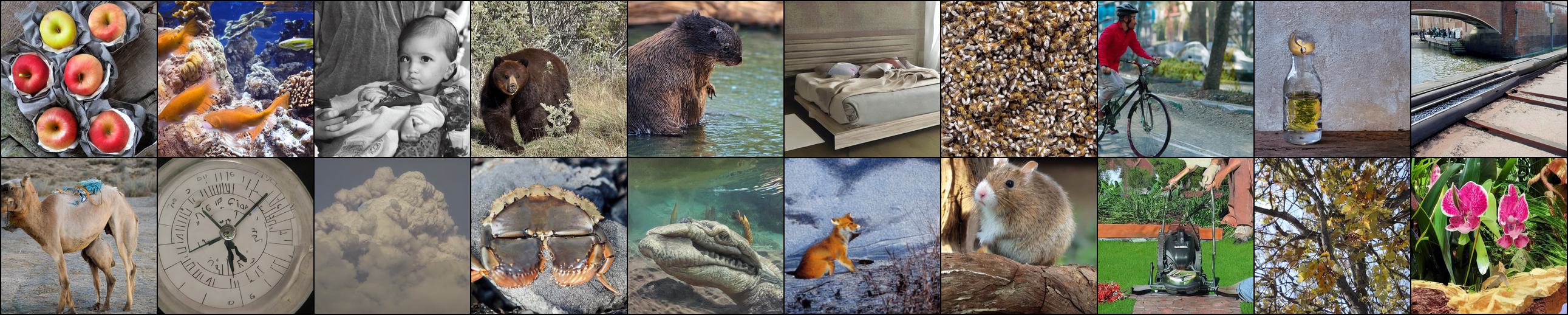}}
\end{center}
   \caption{Samples from $\mathcal{S}$ the set of additional synthetic images generated by Stable Diffusion for class-incremental learning on CIFAR100; images are generated with size 512x512 and are being resized to 32x32 before being fed to the model. Images are displayed before resizing to 32x32 for better appreciation. From left to right, top to bottom: Apple, Aquarium fish, Baby, Bear, Beaver, Bed, Bee, Bicycle, Bottle, Bridge, Camel, Clock, Cloud, Crab, Crocodile, Fox, Hamster, Lawn-mower, Maple tree, Orchids.}
\label{fig:syntheticsamples}
\end{figure*}

In a class-incremental learning (CIL) scenario, deep neural network models are not trained offline on a fixed pre-collected dataset but sequentially on new incoming data. The model has to be incrementally updated using only a limited number of new classes at a time, with the objective of learning a unified classifier among all seen classes. This paradigm, similar to the human learning process, is challenging as models are suffering from catastrophic forgetting~\cite{MCCLOSKEY1989109,ratcliff1990connectionist}: learning new data or classes will result in severe degradation of the performance on the past ones.

Multiple researches have highlighted the importance of rehearsal for continual and incremental learning: by storing samples from each previously encountered class in a memory, models are able to replay past data while learning new ones in order to mitigate catastrophic forgetting. However, the size of this memory is often limited to a few samples per class due to either storage limitation or privacy concerns. Several authors~\cite{lechat2021pseudo,lee2019overcoming,liu2023online,zhang2020class} have proposed to leverage additional external training data as a means to work around this constraint. Those additional training data are sampled from large curated datasets of real images, such as ImageNet, and belong to classes different from the ones encountered by the incremental learner. They are mostly used in an unsupervised manner, for distillating knowledge from the past model to the new one. 

In this work, we propose to use the pretrained state-of-the-art generative model Stable Diffusion~\cite{Rombach_2022_CVPR} as a source of complementary data for class-incremental learning. Compared to approaches that rely on additional external real datasets, our method generates synthetic samples belonging to the same classes as the ones previously encountered by the model. This fundamental difference allows us to make use of those samples not only for the knowledge distillation loss but also for replay in the classification loss whereas concurrent works only focus on the distillation loss. Ablation study highlights the improvement resulting from using the additional data for both losses.
To the best of our knowledge, this is the first attempt to use large pretrained text-to-image generative models to improve general class-incremental learning methods on large scale datasets.

\section{Related work}

\textbf{Class-incremental Learning (CIL):} Methods for class-incremental learning~\cite{zhou2023deep} usually rely on knowledge distillation and a small replay memory containing past samples combined with a bias mitigation method. Knowledge distillation is used to preserve previous knowledge, it may be logit distillation ~\cite{li2017learning,Rebuffi_2017_CVPR,wu2019large}, feature distillation~\cite{douillard2020podnet,Hou_2019_CVPR} or relational distillation~\cite{10.1007/978-3-030-58529-7_16}. Using a memory for replaying previously learned exemplars is a simple yet effective method to recall knowledge of old classes. However, the limited size of the memory induces a bias toward the new classes that strongly degrade the performance of the model if not mitigated. To this end, several methods have been proposed such as Nearest-Mean-of-Exemplars (NME) classifier~\cite{Rebuffi_2017_CVPR}, cosine classifier~\cite{Hou_2019_CVPR}, bias correction layer~\cite{wu2019large}, finetuning on a balanced subset~\cite{10.1007/978-3-030-01258-8_15} or specific losses~\cite{Ahn_2021_ICCV,10.1007/978-3-030-86340-1_31,JODELET2022103582}. To overcome the limitation induced by the limited size of the memory, several works have also proposed to use model inversion~\cite{smith2021always,yin2020dreaming} or to train a generative model in parallel to generate samples from past classes~\cite{he2018exemplar,shin2017continual,wang2021ordisco}. These methods rely on the quality of the generated data, which poses a challenge for large scale datasets and is impacted by catastrophic forgetting when trained sequentially.

\textbf{Additional data for CIL:} Various approaches have studied the use of additional external data, either as a direct substitute in case of the absence of memory~\cite{lee2019overcoming,zhang2020class} or as a complementary resource during training~\cite{lechat2021pseudo,liu2023online}. Global distillation (GD)~\cite{lee2019overcoming} proposed a confidence-based sampling strategy to sample additional external data and used them for a specifically-designed distillation loss. Deep Model Consolidation (DMC)~\cite{zhang2020class} consists in training one distinct model for each training step and use external data to combine them into a unified model. Lechat~\etal~\cite{lechat2021pseudo} proposed a pseudo-labeling approach to better exploit the additional external unlabeled data for representation learning. Liu~\etal~\cite{liu2023online} observed that using samples from new classes in the knowledge distillation loss negatively impacts the training of incremental models and proposed to replace them using placebos of old classes sampled from an external dataset using an online learning method. Those different approaches rely solely on real images, most of the time unlabeled, by sampling their additional data from large curated datasets: primarily ImageNet~\cite{russakovsky2015ImageNet} or TinyImages dataset~\cite{4531741}.

\textbf{Learning from synthetic data:} In one of the pioneer works on learning from synthetic images, Besnier~\etal~\cite{besnier2020dataset} trained a classifier for 10 classes of ImageNet using samples generated by a class-conditional GAN trained on ImageNet. Jahanian~\etal~\cite{jahanian2022generative} used a GAN to generate multiple views for contrastive learning methods. Recently, following the recent breakthrough in text-to-image generative models, Sariyildiz~\etal~\cite{Sariyildiz_2023_CVPR} proposed the use of those general-purpose text-conditioned pretrained generative models for synthesizing datasets as a direct replacement of real image datasets for the training of large scale image-level classification models. While these models can not reach the same accuracy as the models trained using real images when tested on real test datasets, they exhibit other advantages such as strong generalization capability. Concurrently, He~\etal~\cite{he2022synthetic} also proposed to use large synthetic datasets for improving pretrained models on zero-shot and few-shot learning, and transfer learning.

In this work, we propose to study the use of widely available pretrained generative models such as Stable Diffusion~\cite{Rombach_2022_CVPR} as a source of additional labeled samples combinable with general class-incremental learning methods applied on large scale datasets. 

\section{Proposed method}

\begin{figure*}
\begin{center}
\resizebox{\textwidth}{!}{%
\includegraphics{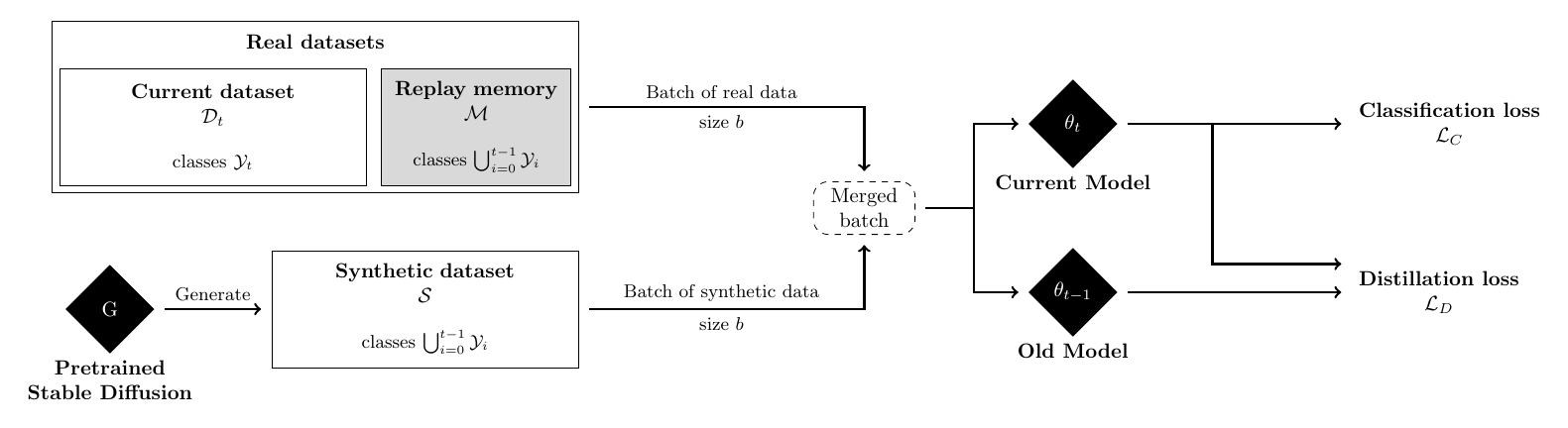}}
\end{center}
   \caption{Diagram representing our proposed method SDDR for leveraging additional external synthetic labeled samples for class-incremental learning. More details in Algorithm~\ref{alg:sddr}.}
\label{fig:diagramsddr}
\end{figure*}

The class-incremental training procedure is divided into $T+1$ steps: the base step, often referred to as step $0$ or initial step, followed by $T$ incremental steps. Each step consists of a training dataset $\mathcal{D}_t$ containing images belonging to the set of new and previously unseen classes $\mathcal{Y}_t$; $\forall i,j \in \{0,1,...,T\}, \mathcal{Y}_i \cap \mathcal{Y}_j = \emptyset$ for $i \neq j$. During each step, the model $\theta_t$ is trained on $\mathcal{D}_t \cup \mathcal{M}$ where $\mathcal{M}$ is the replay memory containing few samples from previously encountered classes. After each step, the model $\theta_t$ is evaluated on the test set of all the classes learned so far without having access to any step or task descriptor. Additionally, during every step, the model also has access to an external source of data $\mathcal{S}$ that may or may not change during the training. This external source of additional data $\mathcal{S}$ can either be an online, and potentially infinite stream of data that the model can fetch when needed or a fixed set of data accessible in an offline manner.

\subsection{Synthetic images}
\label{section:synthetic}

Following the recent in-depth study of Sariyildiz~\etal\cite{Sariyildiz_2023_CVPR}, we propose to use a pretrained Stable Diffusion~\cite{Rombach_2022_CVPR} as the source of additional data $\mathcal{S}$ for incremental learning. 

\textbf{Prompting:} To generate a sample for $\mathcal{S}$, the generative model Stable Diffusion requires a textual prompt. As proposed in~\cite{Sariyildiz_2023_CVPR}, to generate a synthetic image for the class c we use the prompt \enquote{$c, \; d_c$} where \enquote{$c$} is the textual name of the class and \enquote{$d_c$} its description. Although using only \enquote{c} as a prompt may be enough in most cases, this may lead to semantic errors with homographs. For example, in the CIFAR100 dataset using only \enquote{apple} as a prompt may generate images of products or shops from the homonym brand instead of the fruit. Adding the definition of the class to the prompt helps solve this issue. In order to produce these prompts automatically without any further engineering, WordNet~\cite{10.1145/219717.219748} is used to provide the lemmas of the synset as the class name \enquote{c} and the definition of the synset as the description \enquote{$d_c$}. For datasets such as ImageNet, the association of each class with a synset of WordNet is included in the dataset itself. For other datasets such as CIFAR100, the association with WordNet can be done semi-automatically at a negligible cost. Figure~\ref{fig:syntheticsamples} shows synthetic samples for CIFAR100 before being resized to 32x32. In the cases where the mapping between the target dataset and WordNet can not be done, it may be necessary to either manually design and tune the prompt for each class or to rely on a different automation approach such as using image-to-text model for example.

\textbf{Guidance scale:} The classifier-free guidance~\cite{ho2021classifierfree} controls the trade-off between the quality of the generated synthetic samples and their diversity in diffusion models. Following Sariyildiz~\etal~\cite{Sariyildiz_2023_CVPR}, we also used a guidance scale of $2.0$ compared to the default $7.5$ to increase the diversity of our source of additional data at the cost of a slight decrease of the quality of each sample.

\subsection{SDDR for Class-incremental learning}

\begin{algorithm}
\DontPrintSemicolon
\caption{SDDR for class-incremental learning}\label{alg:sddr}
\textbf{Input:} Data-flow $\{\mathcal{D}_i\}_{i=0}^{T}$ and pretrained generative model G. \;
\textbf{Ouput:} Models $\{\theta_i\}_{i=0}^{T}$, replay memory $\mathcal{M}$, and synthetic dataset $\mathcal{S}$. \; \;

\For{$i \in \{0,1,...,T\}$}{
\For{$e \in \{0,1,...,E\}$}{
\While{$(X,Y) \sim \mathcal{D}_i \cup \mathcal{M} $}{
\If{$i > 0$}{
\tcp{If not base step}
$(X_S,Y_S) \sim \mathcal{S}$  \;
$(X,Y) \leftarrow (X\cup X_S, Y\cup Y_S)$ \;
}
Update model $\theta_i$ using $(X,Y)$ \;
}
}
$\mathcal{M} \leftarrow$ UpdateMemory($\mathcal{M}$, $\mathcal{D}_i$) \;
$\mathcal{S} \leftarrow$ UpdateSynthetic($\mathcal{S}$, G, $\mathcal{Y}_i$) \;
}
\end{algorithm}

\begin{table*}
\begin{center}
\begin{tabular}{l ccc c ccc c cc}
\hline
\multirow{ 2}{*}{Methods} & \multicolumn{3}{c}{CIFAR100} & & \multicolumn{3}{c}{ImageNet-Subset} & &  \multicolumn{2}{c}{ImageNet} \\
\cline{2-4} \cline{6-8} \cline{10-11}
& $T$=5 & 10 & 25 & & 5 & 10 & 25 & & 5 & 10 \\
\hline
iCaRL~\cite{Rebuffi_2017_CVPR} & 57.17 & 52.57 & - & & 65.04 & 59.53 & - & & 51.36 & 46.72 \\
BiC~\cite{wu2019large} & 59.36 & 54.20 & 50.00 & & 70.07 & 64.96 & 57.73 & &  62.65 &  58.72 \\
LUCIR~\cite{Hou_2019_CVPR} & 63.42 & 60.18 & - & & 70.47 & 68.09 & - & & 64.34 & 61.28 \\
LUCIR w/ Mnemonics~\cite{liu2020mnemonics} & 63.34 & 62.28 & 60.96 & & 72.58 & 71.37 & 69.74 & & 64.54 & 63.01 \\
LUCIR w/ DDE~\cite{Hu_2021_CVPR} & 65.27 & 62.36 & - & & 72.34 & 70.20 & - & & 67.51 & 65.77 \\
LUCIR w/ BalancedS-CE~\cite{JODELET2022103582}& 64.83 & 62.36 & 58.38 & & 71.18 & 70.66 & 65.10 & & \textbf{67.81} & \textbf{66.47} \\
TPCIL~\cite{10.1007/978-3-030-58529-7_16} & 65.34 & 63.58 & - & & \underline{76.27} & 74.81 & - & & 64.89 & 62.88 \\
DER (w/o Pruning)~\cite{yan2021dynamically} & 68.52 & 67.09 & - & & - & \textbf{78.20} & - & & - & - \\
\hline
iCaRL*~\cite{Rebuffi_2017_CVPR} & 57.68 & 52.44 & 48.05 & &  65.34 & 60.74 & 54.31 & & 53.58 & 48.65 \\
\rowcolor{lightgray!50}
iCaRL w/ SDDR (ours) & 62.01 & 59.24 & 56.47 & & 68.64 & 65.55 & 62.75 & & 57.15 & 54.07 \\
\hline
LUCIR*~\cite{Hou_2019_CVPR} &  63.37 & 60.88 & 57.07 & & 70.75 & 68.48 & 63.14 & & 66.69 & 64.06  \\
\rowcolor{lightgray!50}
LUCIR w/ SDDR (ours) &  65.77 & 63.84 & 61.73 & & 72.92 & 71.48 & 69.48 & & \underline{67.56} & \underline{66.44} \\
\hline
FOSTER*~\cite{wang2022foster} & \underline{71.17} & \underline{68.89} & \underline{65.07} & & 76.09 & 75.05 & \underline{70.96} && - & - \\
\rowcolor{lightgray!50}
FOSTER w/ SDDR (ours) & \textbf{72.18} & \textbf{70.88} & \textbf{68.06} && \textbf{77.13} & \underline{76.77} & \textbf{75.50} && - & - \\
\hline
\end{tabular}
\end{center}
\caption{Average incremental accuracy (Top-1) on CIFAR100, ImageNet-Subset, and ImageNet with a base step containing half of the classes followed by 5, 10, and 25 incremental steps, using a growing memory of 20 exemplars per class. Results for iCaRL and LUCIR are reported from \cite{Hou_2019_CVPR}; results for Mnemonics and BiC are reported from \cite{liu2020mnemonics}. Results for DDE, BalancedS-CE, TPCIL, and DER are reported from their respective paper. Results marked with \enquote{*} correspond to our own experiments. Results on CIFAR100 and ImageNet-Subset are averaged over 3 random runs. Results on ImageNet are reported as a single run. Best result is marked in bold and second best is underlined.}
\label{tab:main}
\end{table*}

Using the previously described generative model, we propose a new method for class-incremental learning: \textbf{S}table \textbf{D}iffusion for \textbf{D}istillation and \textbf{R}eplay (\textbf{SDDR}), which jointly uses the additional synthetic data $\mathcal{S}$ with the real images from the replay memory $\mathcal{M}$.

Datasets of real images used by concurrent approaches as an external source of additional data for class-incremental learning do not contain any image belonging to the same classes as the one encountered by the incremental model. This leads these methods to mainly use this external source of data in an unsupervised manner for knowledge distillation only. By leveraging the pretrained text-to-image generative model, our method is able to generate an additional dataset $\mathcal{S}$ containing labeled images belonging to the same classes as the one previously learned by the model without the need for costly manual labeling and curating process. This allows us to use the additional dataset for distillation but also for replay in the classification loss. The ablation study in Section~\ref{section:ablation} shows the advantages of our approach.

Our approach is designed as a complementary method that can be seamlessly combined with other standard methods for class-incremental learning. Algorithm~\ref{alg:sddr} describes the training procedure of our approach SDDR when combined with a class-incremental learning model. Each step of the class-incremental learning procedure is divided into two distinct phases. During the first phase, the model $\theta_t$ is trained following the method it is combined with, using a mixed batch of data. As illustrated in Figure~\ref{fig:diagramsddr}, each batch used to train the model is composed of half of synthetic images sampled from $\mathcal{S}$ and the other half of real data sampled from the memory and the current dataset $\mathcal{D}_t$. Then in the second phase, before moving to the next incremental step, the synthetic dataset $\mathcal{S}$ is extended by using Stable Diffusion to generate $n$ synthetic samples for each new class encountered in the current dataset $\mathcal{D}_t$, following the method detailed in Section~\ref{section:synthetic}. Therefore, at the end of each incremental step, the additional dataset $\mathcal{S}$ would contain $nN_t$ samples where $N_t =\lvert \bigcup_{i=0}^{t}\mathcal{Y}_i \rvert$ is the number of classes encountered up to the step $t$, included.

In our work, we considered this additional dataset of synthetic samples $\mathcal{S}$ as an offline dataset stored on the device itself. Nonetheless, if the storage is a constraint for the incremental learner, the additional dataset $\mathcal{S}$ can be used in an online way as an infinite stream of data, similarly to~\cite{liu2023online}, by either using the generative model locally or by relying on a cloud-based model. Compared to concurrent works relying on real datasets as a source of additional data, our method offers more flexibility depending on the constraint of the problem: it can be used with or without limited storage, computational budget, and capacity to communicate with external services.

The quality of the synthetic images, their fidelity to the target classes, and the variety of concepts available in the additional external dataset $\mathcal{S}$ are intrinsically limited by the pretrained generative text-to-image model used. Using a different and more specialized generative model or finetuning it on the specific target training dataset may further improve the performance of the incremental learning model it is combined with. 

A naive and straightforward application of a pretrained text-to-image generative model in the context of class-incremental learning would be to use the synthetic datasets $\mathcal{S}$ as a complete replacement of the replay memory for exemplar-free class-incremental learning. However, as discussed in the ablation study in Section~\ref{section:ablation}, this approach does not achieve competitive results.

\section{Experiments}

\begin{table*}
\begin{center}
\begin{tabular}{l ccc c ccc }
\hline
\multirow{ 2}{*}{Methods} & \multicolumn{3}{c}{CIFAR100} & & \multicolumn{3}{c}{ImageNet-Subset} \\
\cline{2-4} \cline{6-8}
& $T$=5 & 10 & 25 & & 5 & 10 & 25 \\
\hline
FOSTER$\dagger$~\cite{wang2022foster} & 70.62 & 68.43 & 63.83 && 80.21 & 77.63 & 69.27 \\
\rowcolor{lightgray!50}
FOSTER w/ PlaceboCIL$\dagger$~\cite{liu2023online} & 71.97 & 70.31 & 67.02 && \textbf{82.03} & \textbf{79.52} & 72.79 \\
\rowcolor{lightgray!50}
\quad \small Improvement in \emph{p.p.}  & \small \textbf{+1.35} & \small +1.88 & \small \textbf{+3.19} && \small \textbf{+1.82} & \small \textbf{+1.89} & \small +3.52 \\
\hline
FOSTER*~\cite{wang2022foster} & 71.17 & 68.89 & 65.07 && 76.09 & 75.05 & 70.96 \\
\rowcolor{lightgray!50}
FOSTER w/ SDDR (ours) & \textbf{72.18} & \textbf{70.88} & \textbf{68.06} && 77.13 & 76.77 & \textbf{75.50} \\
\rowcolor{lightgray!50}
\quad \small Improvement in \emph{p.p.} & \small +1.01 & \small \textbf{+1.99} & \small +2.99 && \small +1.04 & \small +1.72 & \small \textbf{+4.54}  \\
\hline
\end{tabular}
\end{center}
\caption{Average incremental accuracy (Top-1) on CIFAR100, and ImageNet-Subset with a base step containing half of the classes followed by 5, 10, and 25 incremental steps, using a growing memory of 20 exemplars per class. Results marked with \enquote{*} correspond to our own experiments and results marked with \enquote{$\dagger$} are reported from~\cite{liu2023online}. Results averaged over 3 random runs.}
\label{tab:fosterComp}
\end{table*}

\subsection{Experimental setups}

\textbf{Datasets:} Experiments are conducted on three datasets: CIFAR100~\cite{krizhevsky2009learning}, ImageNet-Subset and ImageNet~\cite{russakovsky2015ImageNet}. CIFAR100 is composed of 60,000 32x32 RGB images from 100 classes with 500 training and 100 testing samples per class. ImageNet (ILSVRC 2012) is composed of around 1.28 million high-resolution images from 1,000 classes with about 1,300 training and 50 testing sample per class. ImageNet-Subset is a subset of ImageNet containing only 100 classes. Experiments on an alternative version of ImageNet-Subset can be found in the Supplementary materials. Following the experimental setting initially proposed by Hou~\etal~\cite{Hou_2019_CVPR}, an initial base step containing half of the classes is followed by $T$ (5, 10, or 25) incremental steps containing the remaining classes evenly divided. The class order is defined by NumPy using the random seed 1993. The training images are normalized, randomly horizontally flipped, and cropped, and no more augmentation is applied. For FOSTER~\cite{wang2022foster}, following its authors, AutoAugment~\cite{Cubuk_2019_CVPR} is also used for augmentation.

\textbf{Comparison methods:} We combine our proposed method SDDR with the recent state-of-the-art approach FOSTER~\cite{wang2022foster} and with two fundamental baselines for class-incremental learning: iCaRL~\cite{Rebuffi_2017_CVPR} and LUCIR~\cite{Hou_2019_CVPR}. We selected these two methods because they are representative of the principal approaches for class-incremental learning: using different distillation losses (logits distillation and feature distillation) and difference bias mitigation methods (Nearest-Mean Exemplar classifier and cosine normalized classifier). For comparison, we also include the recent approach PlaceboCIL~\cite{liu2023online} which leverages external unlabeled datasets of real high-resolution images to improve the distillation loss of class-incremental learning methods.

\textbf{Implementation details:} Following the standard setting~\cite{Hou_2019_CVPR,Rebuffi_2017_CVPR}, a 32-layer ResNet~\cite{He_2016} is used for CIFAR100 and a 18-layer ResNet for ImageNet and ImageNet-Subset. Every considered method uses a growing memory containing 20 exemplars per class during the training procedure. When combining SDDR with an existing baseline for class-incremental learning, the same hyperparameters as the original method are used. The only difference is that the batch size during incremental steps is effectively doubled by appending synthetic images generated by the generative model to the original batch of real images from the union of the new dataset and the replay memory. Some methods for class-incremental learning use other losses in addition to the classification and distillation losses. Using synthetic data for these other losses may improve or degrade the performance of the method on a case-by-case basis. Specifically, in our experiments on LUCIR~\cite{Hou_2019_CVPR}, we decide not to use the synthetic data for the margin ranking loss as preliminary experiments suggested that it slightly decreases the performance of the method. Likewise, in our experiments on FOSTER~\cite{wang2022foster}, the synthetic data are used for the feature enhancement loss and the distillation loss during feature boosting and for the distillation loss during feature compression. We do not use the synthetic data for the logits aligned classification loss as it results in a significant decrease of the accuracy due to a bias of the classifier. For fair comparison with the other baselines, we report the accuracy of FOSTER after feature compression. For comparison of dual branch FOSTER before feature compression, see Supplementary materials. The generative model we use is Stable Diffusion version 1.4~\cite{Rombach_2022_CVPR} with a guidance scale of 2.0, 50 steps and the prompt \enquote{$c, \; d_c$}~\cite{Sariyildiz_2023_CVPR}. Synthetic images have been generated using fixed seeds before the incremental training and every method is using the same datasets of synthetic images. The same seeds have been used for generating all classes, resulting in closely related classes having some similar images (same background, subject's pose and position) where only the subject of the image changes. For CIFAR100, the synthetic dataset contains 500 32x32 images per class, and for ImageNet and ImageNet-Subset 1300 512x512 images per class. For every dataset, synthetic images are generated with size 512x512 and are only resized to 32x32 for CIFAR100 prior to the experiments.

\textbf{Performance Measure:} Models are evaluated and compared using the Average Incremental Accuracy defined by Rebuffi~\etal~\cite{Rebuffi_2017_CVPR}. It is the average of the Top-1 accuracy of the model on the test data of all the classes seen so far, at the end of each training step including the initial base step.

\subsection{Comparison with baselines}

\begin{table*}
\begin{center}
\resizebox{\textwidth}{!}{
\begin{tabular}{l ccc c ccc c ccc c ccc}
\hline
\multirow{ 3}{*}{Methods} & \multicolumn{ 15}{c}{CIFAR100} \\
& \multicolumn{3}{c}{5 exemplars/class} & & \multicolumn{3}{c}{10 exemplars/class} & &  \multicolumn{3}{c}{20 exemplars/class} & &  \multicolumn{3}{c}{50 exemplars/class}  \\
\cline{2-4} \cline{6-8} \cline{10-12} \cline{14-16}
& $T$=5 & 10 & 25 & & 5 & 10 & 25 & & 5 & 10 & 25 & & 5 & 10 & 25 \\
\hline
iCaRL*~\cite{Rebuffi_2017_CVPR} & 43.40 & 39.30 & 29.59 & & 52.30 & 46.79 & 38.78 & & 57.68 & 52.44 & 48.05 & & 62.09 & 58.42 & 55.36 \\
iCaRL w/ PlaceboCIL~\cite{liu2023online} & 51.55 & - & - & & 59.11 & - & - & & 61.24 & - & - & & - & - & - \\
\rowcolor{lightgray!50}
iCaRL w/ SDDR (ours) & 55.36 & 50.80 & 48.15 & & 58.53 & 55.70 & 52.51 & & 62.01 & 59.24 & 56.47 & & 65.53 & 63.92 & 62.18\\
\rowcolor{lightgray!50}
\quad \small Improvement in \emph{p.p.} & \small +11.96  & \small +11.50 & \small +18.56 &  & \small +6.23 & \small +8.91 & \small +13.73 &  & \small +4.33 & \small +6.80 & \small +8.42 & & \small +3.44 & \small +5.50 & \small +6.82  \\
\hline
LUCIR*~\cite{Hou_2019_CVPR} &  53.14 & 52.79 & 44.67 & & 60.77 & 57.55 & 50.77 & & 63.37 & 60.88 & 57.07 & & 65.63 & 62.39 & 60.80 \\
LUCIR w/ PlaceboCIL \cite{liu2023online} & 62.74 & - & - & & 64.79 & - & - & & 65.28 & - & - & & - & - & - \\ 
\rowcolor{lightgray!50}
LUCIR w/ SDDR (ours) &  62.90 & 59.81 & 56.32 & & 64.47 & 62.16 & 59.44 & & 65.77 & 63.84 & 61.73 & & 67.21 & 65.81 & 64.48 \\
\rowcolor{lightgray!50}
\quad \small Improvement in \emph{p.p.} & \small +9.76 & \small +7.02 & \small +11.65 &  & \small +3.70 & \small +4.61 & \small +8.67 &  & \small +2.40 & \small +2.96 & \small +4.66 &  & \small +1.58 & \small +3.42 & \small +3.68  \\
\hline
\end{tabular}}
\end{center}
\caption{Average incremental accuracy (Top-1) on CIFAR100 with a base step containing half of the classes followed by 5, 10, and 25 incremental steps depending on the number of exemplars saved in memory for each class. Results marked with \enquote{*} correspond to our own experiments. Results for PlaceboCIL reported from \cite{liu2023online}. Improvement reported comparatively to the baseline method. Results averaged over 3 random runs. }
\label{tab:memory}
\end{table*}

In Table~\ref{tab:main}, the average incremental accuracy (Top-1) of our method and the different baselines are reported for CIFAR100, ImageNet-Subset, and ImageNet with 5, 10, and 25 incremental steps using a replay memory of 20 real exemplars per class for every method. Due to significant differences in the reproduced accuracy for FOSTER~\cite{wang2022foster} between our work and Liu~\etal~\cite{liu2023online}, we decided to make a more detailed comparison with PlaceboCIL in Table~\ref{tab:fosterComp}.

For FOSTER, LUCIR, and iCaRL, using our proposed approach SDDR leads to a significant improvement of the average incremental accuracy: from 1.01 percentage points (\emph{p.p.}) to 4.54\emph{p.p.} for FOSTER, from 0.87\emph{p.p.} to 6.34\emph{p.p.} for LUCIR and from 3.3\emph{p.p.} to 8.44\emph{p.p.} for iCaRL, depending on the dataset and the setting. This improvement is especially important in more challenging settings with numerous incremental steps or when the replay memory is small as discussed in Section~\ref{section:ablation}. When looking more closely at the learned model, we can see that our approach improves the capacity of the model to preserve past knowledge without penalizing its plasticity. This is especially true in the most challenging settings with a small replay memory: for example on CIFAR100 with 10 incremental steps and 5 exemplars per class in the memory, by combining LUCIR with SDDR, the final accuracy on the 50 bases classes increases from 34.15\% to 42.06\% while the accuracy on the 50 remaining classes also increases from 46.08\% to 52.23\%. By combining our proposed method SDDR with FOSTER, we achieve new state-of-the-art performance on several datasets and settings.

Our proposed method achieves competitive performances compared to other considered approaches Mnemonics~\cite{liu2020mnemonics}, DDE~\cite{Hu_2021_CVPR} or BalancedS-CE~\cite{JODELET2022103582} and could be even combined with them to further improve their performance. Compared to PlaceboCIL~\cite{liu2023online}, our method achieves similar or higher performance while using a significantly smaller dataset of additional external data: 50,000 synthetic images compared to the 1.28 million of real images from ImageNet for the experiments on CIFAR100 and 130,000 synthetic images compared to about 1.1 million of real images sampled without overlapping from ImageNet for the experiments on ImageNet-Subset.

\subsection{Ablation Study}
\label{section:ablation}

\begin{table}
\begin{center}
\resizebox{\columnwidth}{!}{
\begin{tabular}{lccc}
\hline
\multirow{ 2}{*}{Methods} & \multicolumn{3}{c}{CIFAR100} \\
\cline{2-4}
 & $T$=5 & 10 & 25 \\
\hline
LUCIR \cite{Hou_2019_CVPR} & 63.37 & 60.88 & 57.07 \\
\quad w/ SD Distillation & 64.62 & 63.12 & \textbf{62.15} \\
\quad w/ SD Distillation w/o new & 62.66 & 61.93 & 61.48 \\
\quad w/ SD Replay & 64.95 & 62.44 & 59.59 \\
\quad w/ SDDR & \textbf{65.77} & \textbf{63.84} & 61.73 \\
\hline
\end{tabular}}
\end{center}
\caption{Average incremental accuracy (Top-1) on CIFAR100 with a base step containing half of the classes followed by 5, 10, and 25 incremental steps, using a growing memory of 20 exemplars per class. Results averaged over 3 random runs.}
\label{tab:component}
\end{table}

\begin{table}
\begin{center}
\begin{tabular}{lccc}
\hline
\multirow{ 2}{*}{Methods} & \multicolumn{3}{c}{CIFAR100} \\
\cline{2-4}
 & $T$=5 & 10 & 25 \\
\hline
LUCIR \cite{Hou_2019_CVPR} & 63.37 & 60.88 & 57.07 \\
\quad w/ SDDR $n=50$ & 64.84 & 62.71 & 60.64 \\
\quad w/ SDDR $n=500$ & 65.77 & 63.84 & 61.73 \\
\quad w/ SDDR $n=1000$ & \textbf{66.41} & 64.91 & 62.73 \\
\quad w/ SDDR $n=2000$ & 66.22 & \textbf{65.03} & \textbf{62.79} \\
\hline
\end{tabular}
\end{center}
\caption{Average incremental accuracy (Top-1) on CIFAR100 with a base step containing half of the classes followed by 5, 10, and 25 incremental steps depending on $n$, the number of synthetic samples generated for each class. Using a growing memory of 20 exemplars per class. Results averaged over 3 random runs.}
\label{tab:diversity}
\end{table}

\textbf{Components:} By leveraging synthetic images supposedly belonging to the same classes as the training dataset, it is possible to use them for both replay and distillation. Table~\ref{tab:component} compares the performance when using the synthetic images only for distillation (w/ SD Distillation), only for replay (w/ SD Replay), and for both combined (w/ SDDR). Distillation and Replay using synthetic data appear to be complementary, Replay achieving higher performances for a short number of incremental steps $T$ and Distillation for a higher number of incremental steps. Combining both in SDDR achieves an overall improvement over each used individually. 

\textbf{Size of the synthetic dataset:} Table~\ref{tab:diversity} shows the impact of the number of synthetic images $n$ generated for each past class used during training. With only 50 synthetic images per class, our method can improve the average incremental accuracy by up to 3.57\emph{p.p.} compared to the standard LUCIR. This can be further increased to 5.72\emph{p.p.} by expanding the size of the synthetic dataset. However, it appears that beyond a certain limit, further increasing the size of the synthetic dataset does not yield any significant improvement. This supports our approach of not using Stable Diffusion in an online manner to generate each training batch if the storage is not a constraint of the system.

\textbf{Size of the replay memory:} Table~\ref{tab:memory} measures the improvement from our method depending on the number of incremental steps and the size of the replay memory containing exemplars of past classes. While the improvement compared to the baseline method is more significant in the case of a small memory size, our method also significantly improves the average incremental accuracy of both LUCIR and iCaRL when combined with a larger replay memory. Likewise, the improvement is more notable for more difficult settings containing numerous incremental steps. For example, when trained for 25 incremental steps with 5 samples per class in the replay memory, SDDR almost doubles the final overall accuracy for iCaRL, increasing it from 20.32\% to 40.39\% and augments the average incremental accuracy by 18.56\emph{p.p.}. Those results highlight that our method can boost the performance of the baseline method it is combined with independently of the setting.  Additional results for CIFAR100 with 5 incremental steps can be found in the Supplementary materials.

\textbf{Use of new data:} In their recent work, Liu~\etal~\cite{liu2023online} emphasized that using data from new classes for the distillation loss was detrimental and proposed to only use data of past classes from the memory in addition to the unlabeled samples from the additional external dataset.  Following their study, we also report in Table~\ref{tab:component} the results of our method while using the distillation loss only on the synthetic images of past classes and on the exemplar from the memory (w/ SD Distillation w/o new). However, it appears that in our case, this significantly decreases the performance of the model. As further discussed below, we suppose that it may be due to the gap between the distribution of synthetic images and the distribution of real images: while real samples from new images are generally harmful to the distillation loss, they may be useful for transferring general knowledge about real data in our case.

\begin{table}
\begin{center}
\resizebox{\columnwidth}{!}{
\begin{tabular}{lccccc}
\hline
\multirow{ 3}{*}{Memory} & \multicolumn{ 5}{c}{CIFAR100} \\
 & \multicolumn{2}{c}{T=5} & & \multicolumn{2}{c}{T=10}\\
\cline{2-3} \cline{5-6}
 & Average & Last & & Average & Last \\
\hline
Real 20 & \textbf{63.37} & \textbf{53.91} & & \textbf{60.88} & \textbf{51.42}\\
Synthetic 20 & 52.38 & 33.68 & & 42.79 & 21.83 \\
Synthetic 100 & 54.54 & 36.56 & & 48.56 & 27.93 \\
Synthetic 500 & 55.77 & 38.78 & & 52.19 & 32.58\\
\hline
\end{tabular}}
\end{center}
\caption{Performances of standard LUCIR~\cite{Hou_2019_CVPR} on CIFAR100 with a base step containing half of the classes followed by 5 and 10 incremental steps depending on the number of exemplars per class saved in the memory and whether the exemplars are real or synthetic images. \enquote{Average} is Average Incremental Accuracy (Top-1) and \enquote{Last} is the final overall accuracy of the model after the last incremental step. the Results averaged over 3 random runs.}
\label{tab:syntheticmem}
\end{table}

\textbf{From synthetic to real images:} As highlighted by Sariyildiz~\etal~\cite{Sariyildiz_2023_CVPR}, while the synthetic images generated by Stable Diffusion tend to be highly similar to real ones, there remains a gap that limits the generalization of the model and may negatively impact the performance. We first observed this phenomenon when using the synthetic dataset generated by Stable Diffusion as a direct replacement for the replay memory in the exemplar-free class-incremental learning setting. As depicted in Table~\ref{tab:syntheticmem}, when using synthetic images as exemplars for the past classes, LUCIR has difficulty generalizing to the real test dataset and achieves significantly lower accuracy compared to using real exemplars in the memory. Moreover, while it performs better than exemplar-based methods used without any replay memory, it is not competitive with approaches specifically designed for this challenging setting. 
Secondly, the impact of the gap between synthetic and real data may also be observed at the classifier level, especially when using our proposed SDDR method in the most challenging settings where the size of the replay memory is highly limited. Fixing old class embeddings learned using real data only, using an NME classifier of real exemplars only, or finetuning the classifier on a small dataset of real images may help in this situation: for example, on CIFAR100 with 25 incremental steps and 5 exemplars per class in the replay memory, using an NME classifier for LUCIR w/ SDDR would improve the average incremental accuracy by 0.89\emph{p.p.}, reaching 57.21\% while for standard LUCIR it would decrease it by 0.01\emph{p.p.}. Sariyildiz~\etal~\cite{Sariyildiz_2023_CVPR} addressed the sim-to-real gap by using strong augmentation policy, and multi-crops and experimentally found that it results in a large performance improvement: 51\% relative increase of the accuracy compared to standard limited augmentation in their simplest variant of synthetic ImageNet-100. The same approach could be used for our method, especially as strong augmentations and larger models have been shown to be beneficial for incremental learning~\cite{lechat2021pseudo}.

\section{Conclusion}
In this work, we showed that pretrained generative text-to-image model can be seamlessly combined with general methods for class-incremental learning in order to further improve their performances. Our approach leverages Stable Diffusion to generate labeled synthetic images belonging to the same classes as the ones previously encountered by the model and use those synthetic data for both the distillation loss and the classification loss. Complete experiments show that our approaches significantly increase the average incremental accuracy of state-of-the-art methods for class-incremental learning, especially in settings with highly restricted memory, and achieve comparable or superior results compared to other competitive approaches relying on additional real data samples while offering more flexibility.

While our work focused on a simple and straightforward integration of the synthetic images into the class-incremental learning setting, it would be important in future works to consider how the generative models could be finetuned during the training and how the different losses could be modified to better take advantage of the synthetic images and reduce the synthetic-to-real gap. Moreover, we will also explore the different possible uses of those pretrained generative models for class-incremental learning: for example to prepare the model by learning in advance, during the initial step, classes that may be encountered in the future, based on already encountered classes.

\section*{Acknowledgments}
This work was supported by JSPS Grant-in-Aid for Scientific Research (Grant Number 23H03451, 21K12042).
{\small
\bibliographystyle{ieee_fullname}
\bibliography{egbib}
}

\clearpage

\setcounter{table}{0}
\renewcommand{\thetable}{S\arabic{table}}
\setcounter{figure}{0}
\renewcommand{\thefigure}{S\arabic{figure}}
\setcounter{section}{0}

\noindent
{\Large {\textbf{Supplementary materials}}}
\\

\section{Alternative version of ImageNet-Subset}
Additionally, we considered an alternative set of classes for ImageNet-Subset, denoted as \enquote{ImageNet-Subset Alt.} in this work, used by the authors of FOSTER~\cite{wang2022foster}. In Table~\ref{tab:imagenetalt}, we report the performance of FOSTER combined with our method on this variation of ImageNet-Subset.

Compared to ImageNet-Subset, our proposed method SDDR achieves both higher average incremental accuracy and higher improvement over the baseline method on ImageNet-Subset Alt. .

\begin{table}[h]
\begin{center}
\resizebox{\columnwidth}{!}{
\begin{tabular}{lccc}
\hline
\multirow{ 2}{*}{Methods} & \multicolumn{3}{c}{ImageNet-Subset Alt.} \\
\cline{2-4}
 & $T$=5 & 10 & 25 \\
 \hline
FOSTER* & 78.52 & 76.49 & 71.24\\
\rowcolor{lightgray!50}
FOSTER w/ SDDR (ours) & 80.35 & 79.21 & 77.09 \\
\rowcolor{lightgray!50}
\quad \small Improvement in \emph{p.p.} & \small +1.83 & \small +2.72 & \small +5.85 \\ 
\hline
\end{tabular}}
\end{center}
\caption{Average incremental accuracy (Top-1) on ImageNet-Subset Alt. with a base step containing half of the classes followed by 5, 10, and 25 incremental steps. \enquote{ImageNet-Subset Alt.} is a different version of ImageNet-Subset following the definition of the authors of FOSTER~\cite{wang2022foster}. Results marked with \enquote{*} correspond to our own experiments. Results averaged over 3 random runs.}
\label{tab:imagenetalt}
\end{table}

\section{Dual branch FOSTER}
In our main experiments, to fairly compare our proposed method with other methods, we only evaluated the accuracy of FOSTER after performing the feature compression. To better compare with expansion-based method such as DER~\cite{yan2021dynamically} whose number of parameters is growing at each incremental steps,  we report in Table~\ref{tab:b4compression} the accuracy of the dual branch FOSTER model before the compression, denoted FOSTER-B4.

Experiments show that, if the memory and computational budget of the system allow it, using the dual branch FOSTER for inference with our method further improve the accuracy. On CIFAR100, FOSTER combined with SDDR achieves higher performances than DER for both single and dual branch evaluation. Furthermore, on ImageNet-Subset with 10 incremental steps, FOSTER-B4 combined with SDDR achieves a Top-1 Average incremental accuracy  0.49\emph{p.p.} lower and a Top-5 Average incremental accuracy 0.50\emph{p.p.} higher than DER while requiring about five times less parameters.

\begin{table*}
\begin{center}
\begin{tabular}{l ccc c ccc}
\hline
\multirow{ 2}{*}{Methods} & \multicolumn{3}{c}{CIFAR100} & & \multicolumn{3}{c}{ImageNet-Subset} \\
\cline{2-4} \cline{6-8}
& $T$=5 & 10 & 25 & & 5 & 10 & 25 \\
\hline
DER (w/o Pruning)~\cite{yan2021dynamically} & 68.52 & 67.09 & - & & - & \textbf{78.20} & - \\
\hline
FOSTER*~\cite{wang2022foster} & 71.17 & 68.89 & 65.07 & & 76.09 & 75.05 &70.96 \\
\rowcolor{lightgray!50}
FOSTER w/ SDDR (ours) & \underline{72.18} & \underline{70.88} & \underline{68.06} && 77.13 & 76.77 & \underline{75.50}  \\
\hline
FOSTER-B4*~\cite{wang2022foster} & 72.08 & 69.32 & 65.41 & & \underline{77.48} & 76.11 & 71.40 \\
\rowcolor{lightgray!50}
FOSTER-B4 w/ SDDR (ours) & \textbf{73.28} & \textbf{71.43} & \textbf{68.36} && \textbf{78.48} & \underline{77.71} & \textbf{75.97} \\
\hline
\end{tabular}
\end{center}
\caption{Average incremental accuracy (Top-1) on CIFAR100, and ImageNet-Subset with a base step containing half of the classes followed by 5, 10, and 25 incremental steps, using a growing memory of 20 exemplars per class. Results for DER are reported from \cite{yan2021dynamically}. Results marked with \enquote{*} correspond to our own experiments. Results on CIFAR100 and ImageNet-Subset are averaged over 3 random runs. Best result is marked in bold and second best is underlined.}
\label{tab:b4compression}
\end{table*}

\section{CIFAR-100 with 5 incremental steps}

Following Liu~\etal~\cite{liu2023online}, we report in Table~\ref{tab:memorySUPP} additional results for the different baseline methods on CIFAR100 with a base step containing half of the classes followed by 5 incremental steps depending on the number of exemplars saved in memory for each class.

When FOSTER~\cite{wang2022foster} is used with a really small replay memory, we observed that the improvement resulting from using our proposed method SDDR is limited. We suppose that this is due to the logits alignment loss used by the authors that too strongly limits the plasticity of the model and restricts our proposed method. By tuning the hyperparameters of the logits aligment loss, it should be possible to further improve the performance of FOSTER when combined with our proposed method.

\begin{table*}
\begin{center}
\resizebox{\textwidth}{!}{
\begin{tabular}{l cc c cc c cc c cc}
\hline
\multirow{ 3}{*}{Methods} & \multicolumn{ 11}{c}{CIFAR100 with $T$=5} \\
& \multicolumn{2}{c}{5 exemplars/class} & & \multicolumn{2}{c}{10 exemplars/class} & &  \multicolumn{2}{c}{20 exemplars/class} & &  \multicolumn{2}{c}{50 exemplars/class}  \\
\cline{2-3} \cline{5-6} \cline{8-9} \cline{11-12}
& Average & Last && Average & Last && Average & Last && Average & Last \\
\hline
iCaRL*~\cite{Rebuffi_2017_CVPR} &  43.40 & 31.90 && 52.30 & 40.11 && 57.68 & 47.45 &&  62.09 & 54.23 \\
\quad w/ PlaceboCIL~\cite{liu2023online} & 51.55 & 39.35 && 59.11 & 46.42 && 61.24 & 51.47 && - & - \\ 
\rowcolor{lightgray!50}
\quad w/ SDDR (ours) & 55.36 & 43.13 && 58.53 & 48.33 && 62.01 & 52.91 && 65.53 & 58.46 \\
\hline
LUCIR*~\cite{Hou_2019_CVPR} & 53.14 & 41.52 && 60.77 & 49.65 && 63.37 & 53.91 && 65.63 & 57.58 \\
\quad w/ PlaceboCIL~\cite{liu2023online} & 62.74 & 53.25 && 64.79 & 55.44 && 65.28 & 56.23 && - & -  \\ 
\rowcolor{lightgray!50}
\quad w/ SDDR (ours) & 62.90 & 50.94 && 64.47 & 54.53 && 65.77 & 57.00 && 67.21 & 59.73 \\
\hline
FOSTER*~\cite{wang2022foster} & 56.94 & 43.74 && 63.09 & 54.34 && 71.17 & 63.97 && 70.00 & 64.10 \\
\quad w/ PlaceboCIL~\cite{liu2023online} & 62.78 & 50.72 && 65.12 & 54.81 && 71.97 & 64.43 && - & -  \\ 
\rowcolor{lightgray!50}
\quad w/ SDDR (ours) & 57.76 & 44.32 && 63.26 & 54.32 && 72.18 & 64.78 && 71.66 & 66.04 \\
\hline
\end{tabular}}
\end{center}
\caption{Performances on CIFAR100 with a base step containing half of the classes followed by 5 incremental steps depending on the number of exemplars saved in memory for each class. Results for PlaceboCIL reported from \cite{liu2023online}. Results marked with \enquote{*} correspond to our own experiments. \enquote{Average} is Average Incremental Accuracy (Top-1) and \enquote{Last} is the final overall accuracy of the model after the last incremental step. Results averaged over 3 random runs. }
\label{tab:memorySUPP}
\end{table*}

\end{document}